\documentclass{sig-alternate-br}
\usepackage{footnote}

\begin{document}

\conferenceinfo{32$^{nd}$ Twente Student Conference on IT}{Jan. 31$^{st}$, 2019, Enschede, The Netherlands.}
\CopyrightYear{2019} 

\title{Automated Machine Learning Techniques \\ for Data Streams}

%
%
%
%
%

\numberofauthors{1} 
%
\author{
%
%
\alignauthor
Alexandru-Ionut Imbrea\\
       \affaddr{University of Twente}\\
       \affaddr{P.O. Box 217, 7500AE Enschede}\\
       \affaddr{The Netherlands}\\
       \email{a.imbrea@student.utwente.nl}
}

\maketitle

\begin{abstract}
Automated Machine Learning (AutoML) techniques benefitted from tremendous research progress recently. These developments and the continuous-growing demand for machine learning experts led to the development of numerous AutoML tools. Industry applications of machine learning on streaming data become more popular due to the increasing adoption of real-time streaming in IoT, microservices architectures, web analytics, and other fields. However, the AutoML tools assume that the entire training dataset is available upfront and that the underlying data distribution does not change over time. These assumptions do not hold in a data-stream-mining setting where an unbounded stream of data cannot be stored and is likely to manifest concept drift. This research surveys the state-of-the-art open-source AutoML tools, applies them to real and synthetic streamed data, and measures how their performance changes over time. For comparative purposes, batch, batch incremental and instance incremental estimators are applied and compared. Moreover, a meta-learning technique for online algorithm selection based on meta-feature extraction is proposed and compared, while model replacement and continual AutoML techniques are discussed. The results show that off-the-shelf AutoML tools can provide satisfactory results but in the presence of concept drift, detection or adaptation techniques have to be applied to maintain the predictive accuracy over time.
\end{abstract}



\section{Introduction}
Developing machine learning models that provide constant high predictive accuracy is a difficult task that usually requires the expertise of a data scientist. Data scientists are multidisciplinary individuals possessing skills from the intersection of mathematics, computer science, and business/domain knowledge. Their job mainly consists of performing a workflow that includes, among others, the following steps: data gathering, data cleaning, feature extraction, algorithm selection, and hyperparameter optimization. The last three steps of this workflow are iterative tasks that involve fine-tuning, which is usually performed by data scientists in a trial-and-error process until the desired performance is achieved. The ever-growing number of machine learning algorithms and hyperparameters leads to an increase in the number of configurations which makes data scientists' job more laborious than ever. 

Considering the above reason and due to the lack of experts required in the industry, the field of Automated Machine Learning (AutoML) benefited from considerable advances recently \cite{elshawi2019automated}. AutoML tools and techniques enable non-experts to achieve satisfactory results and experts to automate and optimize their tasks. Although the field of AutoML is relatively new, it consists of multiple other sub-fields such as Automated Data Cleaning (Auto Clean), Automated Feature Engineering (Auto FE), Hyperparameter Optimization (HPO), Neural Architecture Search (NAS), and Meta-Learning \cite{zoller2019survey}. The sub-field of Meta-Learning is concerned with solving the algorithm selection problem \cite{rice1976algorithm} applied to ML by selecting the algorithm that provides the best predictive performance for a data set. HPO techniques are used to determine the optimal set of hyperparameters for a learning algorithm, while Auto FE aims to extract and select features automatically. NAS represents the process of automating architecture engineering for artificial neural networks which already outperformed manually designed ones in specific tasks such as image classification \cite{elsken2018neural}. The overarching goal of the field is to develop tools that can produce end-to-end machine learning pipelines with minimal intervention and knowledge. However, this is not yet possible using state-of-the-art open-source tools, thus when talking about AutoML throughout the paper it means solving the \textit{combined algorithm selection and hyperparameter optimization} problem, or CASH for short, as defined by Thornton et al. in \cite{thornton2013autoweka}. When other techniques from the general field of AutoML, such as meta-learning, are needed, it is specifically mentioned.

Regardless of their increasing popularity and advances, current AutoML tools lack applicability in data stream mining. Mainly due to the searching and optimization techniques these tools use internally, as described in Section \ref{relatedwork}, they have to make a series of assumptions \cite{madrid2019towards}. First, the entire training data has to be available at the beginning and during the training process. Second, it is assumed that the data used for prediction follows the same distribution as the training data and it does not change over time. However, streaming data may not follow any of these assumptions: it is an unbounded stream of data that cannot be stored entirely in memory and that can change its underlying distribution over time. The problem of applying AutoML to data streams becomes more relevant if considering that the popularity of microservices and event-driven architectures is also increasing considerably \cite{richards2015microservices}. In these types of systems, streams of data are constantly generated from various sources including sensor data, network traffic, and user interaction events. The growing amount of streamed data pushed the development of new technologies and architectures that can accommodate big amounts of streaming data in scalable and distributed ways (e.g. Apache Kafka \cite{kreps2011kafka}). Nevertheless, despite their relevance, AutoML tools and techniques lack integration with such big data streaming technologies.

The workaround solution adopted by the industry consists, in general, of storing the data stream events in a distributed file system, perform AutoML on that data in batch, serialize the model, and use the model for providing real-time predictions for the new stream events \cite{uber2017michelangelo}, \cite{gctables}. This solution presents a series of shortcomings including predicting on an outdated model, expensive disk and network IO, and the problem of not adapting to concept drift. In this paper, the problem of concept drift affecting the predictive performance of models over time is extensively discussed and measured for models generated using AutoML tools. For comparison, batch, batch incremental and online models are developed and used as a baseline. Moreover, detection and adaptation techniques for concept drift are assessed for both AutoML-generated and online models. 

The main contribution of this research consists of:
\begin{itemize}
  \item An overview and discussion of the possible AutoML techniques and tools that can be applied to data streams.
  \item A Python library called \textbf{automl-streams}\footnote{\url{https://pypi.org/project/automl-streams}} available on GitHub\footnote{\url{https://github.com/AlexImb/automl-streams}} including: a way to use a Kafka stream in Python ML/AutoML libraries, an evaluator for pretrained models, implementations of meta-learning algorithms.
  \item A collection of containerized experiments\footnote{\url{https://github.com/AlexImb/automl-streams/tree/master/demos}} that can be easily extended and adapted to new algorithms and frameworks.
  \item An interpretation of the experimental results.
 
\end{itemize}

In the following sections of the paper the main research questions are presented, the formal problem formulations are given for the concepts intuitively described above, and the related work is summarised. Finally, the experimental methods and the results are described.

\section{Research Questions}
The proposed research aims to provide relevant experimental results and interpretations of the results as answers for the following research questions: 

\textbf{RQ1} How can AutoML techniques and tools be applied to data streams?
\newline
\newline
\textbf{RQ2} How do AutoML-tuned models perform over time compared to offline and online models, in providing real-time predictions?
		\newline\-\hspace{0.5cm}\textbf{RQ2.1} How does algorithm selection using online meta-learning influence the predictive performance?
    
\section{Problem formulation}
The formal definition of the AutoML problem as stated by Feurer et al. in \cite{feurer2015autosklearn} is the following:

\textbf{Definition 1 (AutoML):} 
\\ For $i = 1, \dots , n + m$, $n,m \in \mathbb{N}^+$,  let $x_i \in \mathbb{R}^d$ denote a feature vector of $d$ dimensions and $y_i \in Y$ the corresponding target value. Given a training dataset $D_{train} = {(x_1, y_1),\dots , (x_{n}, y_{n})}$ and
the feature vectors $x_{n+1}, . . . , x_{n+m}$ of a test dataset $D_{test} = {(x_{n+1}, y_{n+1}), \dots ,(x_{n+m}, y_{n+m})}$
drawn from \textit{the same underlying data distribution}, as well as a resource budget $b$ and a loss metric
$\mathcal{L}(\cdot, \cdot)$, the AutoML problem is to (automatically) produce test set predictions $y_{n+1}, . . . , y_{n+m}$. The
loss of a solution $\hat{y}_{n+1}, . . . , \hat{y}_{n+m}$ to the AutoML problem is given by:

\begin{equation}
   \frac{1}{m} \sum_{j=1}^{m} \mathcal{L}( \hat{y}_{n+j},  {y}_{n+j})
\end{equation}

When restricting the problem to a combined algorithm selection and hyperparameter optimization problem (CASH) as defined and formalised by Thornton et al. in \cite{thornton2013autoweka} the definition of the problem becomes:

\textbf{Definition 2 (CASH):}
\\ Given a set of algorithms $\mathcal{A} = \{A^{(1)}, \dots, A^{(k)}\}$ with associated hyperparameter domains $\vLambda^{(1)}, \dots, \vLambda^{(k)}$ and a loss metric $\mathcal{L}(\cdot, \cdot, \cdot)$, we define the CASH problem as computing:

\begin{equation}
{A^*}_{\vlambda^*} \in \argmin_{A^{(j)} \in \mathcal{A}, \vlambda \in \vLambda^{(j)}} \frac{1}{k}  \sum_{i=1}^{k} \mathcal{L}(A^{(j)}_\vlambda, \mathcal{D}_{\text{train}}^{(i)}, \mathcal{D}_{\text{valid}}^{(i)})
\end{equation}

\vspace{0.25cm}
\section{Background}

\subsection{AutoML Frameworks}
To solve the AutoML problem (in its original form or as a CASH problem) a configuration space containing the possible combinations of algorithms and hyperparameters is defined. In the case of artificial neural networks algorithms, an additional dimension represented by the neural architecture is added to the search space. For searching this space, different searching strategies and techniques can be used \cite{truong2019towards}. For the purpose of this research one representative open-source framework \cite{gijsbers2019open} was selected for each type of searching strategy (CASH solver). The selected frameworks and their searching strategy are displayed in Table \ref{table:libraries}.

\begin{savenotes}
\begin{table}[h]
\renewcommand{\arraystretch}{1.25}
\centering
\begin{tabular}{|l|l|}
\hline
\textbf{AutoML Framework} & \textbf{CASH Solver} \\ \hline
AutoWeka\footnote{\url{https://www.automl.org/automl/autoweka}} & Bayesian Optimization \\ \hline
H2O.ai\footnote{\url{https://www.h2o.ai/products/h2o}} & Grid Search \\ \hline
TPOT\footnote{\url{http://epistasislab.github.io/tpot}} & Genetic Programming \\ \hline
auto-sklearn\footnote{\url{https://automl.github.io/auto-sklearn/master}} &  SMAC \\ \hline
\end{tabular}
\caption{Open-source AutoML Libraries}
\label{table:libraries}
\end{table}
\end{savenotes}

\textbf{Bayesian Optimization} is an iterative optimization process suited for expensive objective functions. It consists of two main components: surrogate models for modelling the objective function, and an acquisition function that measures the value that would be generated by the evaluation of the objective function at a new point \cite{zoller2019survey}. This technique is also used in the Sequential Model-based Algorithm Configuration (SMAC) library that allows using Gaussian processes and Random Forests as surrogate models \cite{feurer2015autosklearn}.

\textbf{Grid Search}, as the name suggests, creates a grid of configurations and searches through them. The advantage of this approach is that it can be easily parallelized. The H2O.ai framework makes use of that in order to solve the CASH problem in a distributed way across nodes in a cluster \cite{h2o}. 

\textbf{Genetic Programming} is a technique inspired by the process of natural selection where concepts such as chromosomes and mutations are used to develop better generations of solutions to the problem. Tree-based Pipeline Optimization Tool (TPOT) uses this technique to generate and optimize machine learning pipelines \cite{tpot}.

\subsection{Online learning}
Online machine learning approaches such as instance incremental or batch incremental learning are techniques usually applied to data stream mining and big data. These algorithms do not require to store the entire training dataset in memory and can dynamically adapt to changes in the data distribution. While some algorithms are specially designed for online learning \cite{bifet2012ensembles} others are an adaptation of batch algorithms such as Stochastic Gradient Descent, k Nearest Neighbour and Naive Bayes \cite{van2014algorithm}.

\textbf{Ensemble} techniques train a homogeneous \cite{bifet2012ensembles} or heterogeneous \cite{van2018online} set of estimators generating a set of models. In order to make predictions, a voting method between the members of the ensemble is established. Each model makes a prediction and the final prediction is calculated based on a predefined rule such as the majority vote, weight, etc.

\subsection{Concept drift}

Concept drift represents the problem of streamed data not following an underlying distribution and the concept being learned to change over time. Consequently, predictions made by models become less accurate as time passes. While some algorithms such as k Nearest Neighbour can naturally deal with drift by design \cite{van2016massively} others need explicit drift detection methods. All drift detectors assume that the accuracy of a predictor over a data stream should improve or stay the same over time. When the accuracy drops (abruptly or gradually) a warning level or a change is identified. A simple solution is the Page Hinkley test proposed in 1954 \cite{page1954continuous} that involves computing the mean of the observed values up to the current moment. More recent solutions adapted to data stream mining consist of keeping a sliding window and computing relevant statistics over that window. Fixed-size slide window implementations are used for the Drift Detection Method (DDM) \cite{gama2004learning} and Early Drift Detection Method (EDDM) \cite{baena2006early}. The difference between these two is that EDDM aims to improve the detection rate of gradual concept drift in DDM while keeping a good performance against abrupt concept drift. Another popular implementation of a drift detector is ADaptive WINdowing (ADWIN) \cite{bifet2007learning} which will decide the size of the window by cutting the statistics window at different points and analysing the average of some statistic over these two windows.

\vspace{0.5cm}

\subsection{Evaluation techniques}
\label{evaluation}

Considering the non-stationary nature of an unbounded stream of data, classic techniques for evaluating the model on batch data such as train-test split and cross-validation do not apply to models trained on streamed data \cite{gama2009issues}. To overcome this problem and obtain accurate measurements over time other evaluation methods are introduced.

\textbf{Periodic holdout} evaluation involves storing a predefined number of samples from the stream (holdout) for testing purposes and renewing the test samples after a preset amount of time or observed samples. This technique implies that the data used for testing is never used for training the model.

\textbf{Prequential} evaluation or the interleaved test-then-train method in which each sample serves two purposes: first the sample is used for testing, by making a prediction and updating the metrics, and then used for training (partially fitting) the model. Using this method, all the samples are used for training and no holdout has to be kept in memory.

\subsection{Meta-learning}

Meta-learning, or learning how to learn \cite{vanschoren2018meta}, although way older than AutoML, is a field of machine learning that is often categorized as sub-field of AutoML. Inspired by how humans learn and the fact that they do not start from scratch every time but use their previous learning experience to learn new tasks, meta-learning tries to achieve the same for machine learning tasks. Meta-learning techniques can be used for either improving existing models, generating new ones or reducing the configuration space. To do so in a systematic and data-driven way, the collection of meta-data is required. The source of meta-data can differ between the methods used: configurations and performance from model evaluation, task properties such as meta-features, and prior models (transfer learning) \cite{vanschoren2014openml}.

When applied to data streams, meta-learning techniques are often used to predict the weights of base-learners trained on the incoming samples as part of an ensemble \cite{van2018online}. The model that predicts the weights of the base-learners or one base learner (i.e. 100\% weight) is usually called a \emph{meta-learner}. The source of its knowledge (meta-knowledge) can be meta-data represented by meta-features \cite{rossi2017guidance} extracted from the current stream or similar streams following the same concept. Furthermore, the meta-learner can be trained during the processing of the stream or pre-trained and only used for making online predictions during the processing part.

\section{Related Work}
\label{relatedwork}

From the literature research carried, it became clear that the state-of-the-art tools and techniques do not include any solution for solving the AutoML or CASH problem in a streaming setting. However, being a well-known problem in the field, it was part of the NIPS 2018 AutoML Challenge\footnote{\url{https://www.4paradigm.com/competition/nips2018}} formulated as a \emph{life-long AutoML problem}. According to the organisers of the challenge, what is considered to be the main difference between life-long machine learning (LML) and online learning is that in LML the true labels can arrive days or weeks later. Either way, the problem of solving the AutoML on a data stream still holds. Two solutions that performed well in the challenge are going to be discussed. 

First, Wilson et al. propose AutoGBT \cite{wilson2020automatically}, a solution that combines an adaptive self-optimized end-to-end machine learning pipeline. It is based on Gradient Boosting Decision Trees (GBDT) with automatic hyper-parameter tuning using Sequential Model-Based Optimization (SMBO-TPE). Their solution does not include any explicit drift detection mechanism and relies on the implicit drift adaptation techniques of the implementation of GBDT from the LightGBT\footnote{\url{https://github.com/microsoft/LightGBM}} library. Although their solution performed very well for most of the datasets of the challenge, the usage of a single type of algorithm can be seen as a happy path by not being flexible for other tasks or use cases.

Second, Madrid et al. propose LML auto-sklearn \cite{madrid2019towards}, a solution built around the auto-sklearn library, which incorporates explicit drift detection using Fast Hoeffding Drift Detection Method (FHDDM). When drift is detected a decision of either replacing or improving the model is made. According to their benchmarks, the best results are obtained when the model is replaced by retraining a new one with the entire dataset seen so far. However, this method works only until the dataset kept in memory reaches a certain size which is not a feasible solution for big data.

Furthermore, in other work, aspects of sub-fields of AutoML are applied to data streams and online learning. An example will be meta-learning, used for algorithm selection based on meta-features \cite{rossi2017guidance}. Two possible implementations of meta-learning algorithms for streams that can suggest the best predictor for the next sliding window are extensively discussed in the literature. First, Rossi et al. proposed MetaStream \cite{rossi2014metastream}, a meta-learning based method for periodic algorithm selection between two algorithms. The approach of MetaStream is to characterize, i.e. extract meta-features, from a training window and predict the learning algorithm on a selection window consisting of future data points. This way, both characteristics from the past and incoming data are used in the selection process. Second, van Rijn et al. \cite{van2014algorithm} proposed a slightly different approach that involves determining which algorithm among multiple ones will be used to predict the next window of data based on data characteristics measured in the previous window and the meta-knowledge. Both approaches claim to perform better than incremental learning algorithms and ensembles for their selected datasets.

Consequently, after claiming that extracting meta-features is a computationally expensive process, van Rijn et al. \cite{van2018online} proposes a different approach by introducing the Online Performance Estimation framework. It represents a measurement of how ensemble members have performed on recent examples and adjust their weight in the voting accordingly. Another solution, Best-last (BLAST), can select an active (leader) estimator based on the Online Performance Estimation by choosing the one that performed best over the set of w previous training examples.

\section{Methods}
\label{methods}

\begin{figure}
\centering 
\epsfig{file=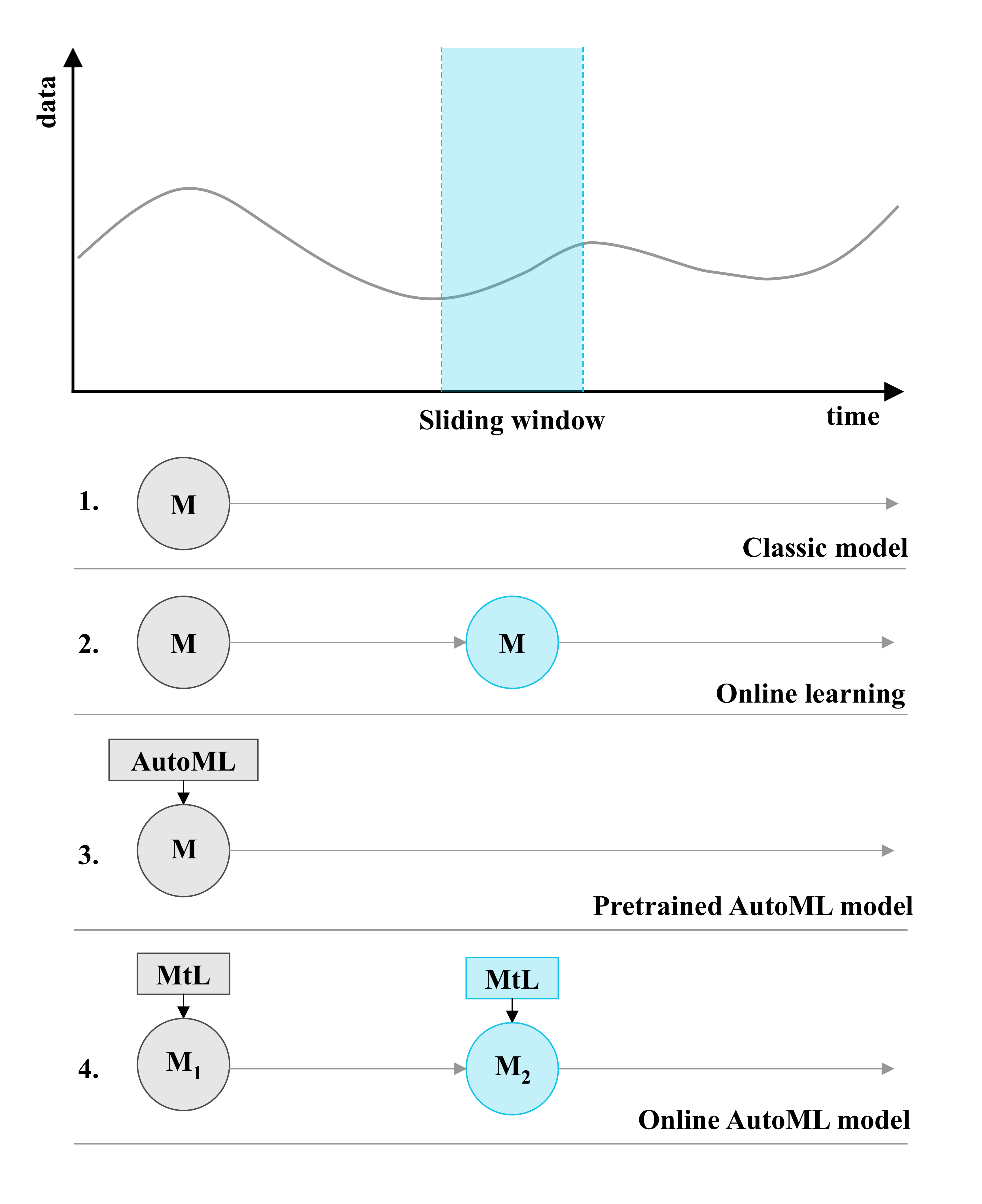, width=8.5cm}
\caption{Experiment Types}
\label{fig:experiments}
\end{figure}

A visual representation of the proposed experiment types can be observed in Figure \ref{fig:experiments}. A stream of data that can experience concept drift over time is depicted at the top of the figure and four experiment types following the same time axis under it. These experiment types are an abstraction of the concrete experiments described in Section \ref{experiments} and aim to provide a higher-level overview of the methods experimentally implemented.

\textbf{Classic model.} The first experiment type is represented by a model fitted with data a priori. The batch of data used for training is collected from the stream and it is ensured that it is not streamed later on (i.e. (parts of) the training data never used for testing). The types of algorithms used in this case are classic batch estimators such as NaiveBayes, LogisticRegression and Linear SVC.

\textbf{Online learning.} The second experiment type implies using online learning algorithms than can perform (batch) incremental learning or an ensemble of those such as Hoeffding Tree, OzaBag, OzaBoost and LeverageBagging. Some of these algorithms might include drift adaptation or detection techniques such as ADWIN or use a window for training weak-learners on the batch data collected in that window.  

\textbf{Pretrained AutoML model.} While the first two types of experiments do not involve any AutoML approach and serve as a baseline for the other experiments in this research, the third experiment consists of using state-of-the-art AutoML libraries to generate the pipeline, select the algorithms and their hyperparameters. 

\textbf{Online AutoML model.} The last experiment type involves implementing theoretically described meta-learning algorithms such as BLAST or MetaStream and ensembles that use meta-learning internally. In contrast with the second experiment type, this one implies the existence of a meta-learner (MtL) that determines which model is used for making predictions or the weight of a model in ensembles. Therefore, at a certain point in time or for a certain time window, the meta-learned can select a different model $M_{1,2, ... n}$ while in the case of the second experiment the model $M$ is the only model making a prediction but evolves over time.

The benchmark data sets include both real and synthetic ones that are used in the related work and in literature in general \cite{van2016massively}. In order to provide verifiability and reproducibility all the datasets will be published on OpenML\footnote{\url{https://openml.org/}} under an OpenML Study (a collection of datasets) called \textit{AutoML Streams}. An overview of the datasets and their characteristics is provided in Table \ref{table:datasets}. While the techniques and methods presented in this paper can be extended to multi-label classification and regression tasks, to reduce the scope of the experiments, only single-label multi-class classification problems were selected.   

\begin{table}[h]
\centering
\renewcommand{\arraystretch}{1.25}
\begin{tabular}{|c|c|c|c|c|} \hline
\textbf{Dataset Name} & \textbf{Type} & \textbf{$|x_n|$} & \textbf{$classes$} & \textbf{$|cls.|$} \\ \hline
agrawal\_gen & generated & 9 & $\{0, 1\}$ & 2 \\ \hline
stagger\_gen & generated & 3 & $\{0, 1\}$ & 2 \\ \hline
sea\_gen & generated & 3 & $\{0, 1\}$ & 2 \\ \hline
led\_gen & generated & 24 & $\{0,\dots,9\}$ & 10 \\ \hline
hyperplane\_gen & generated & 10 & $\{0, 1\}$ & 2\\ \hline
rbf\_gen & generated & 10 & $\{0, 1\}$ & 2\\ \hline
covtype & real & 55 & $\{1,\dots,7\}$ & 7 \\ \hline
elec & real & 6 & $\{0, 1\}$ & 2 \\ \hline
pokerhand & real & 10 & $\{0,\dots,5\}$ & 6 \\ \hline
\end{tabular}
\caption{Datasets used for experiments}
\label{table:datasets}
\end{table}

\textbf{Agrawal Generator} was introduced by Agrawal et al. in \cite{agrawal1993database}. It represents a common source of data for early work on scaling up decision tree learners. The generator produces a stream containing nine features, six numeric and three categorical. There are ten functions defined for generating binary class labels from the features. Presumably, these determine whether a loan should be approved or not. Concept drift is introduced by changing the label-generating function.

\textbf{Stagger Generator} generates a data stream with abrupt concept drift, as described in Gama, Joao, et al. in \cite{gama2004learning}. Each instance describes the size, shape and colour of such an object. A STAGGER concept is a binary classification rule distinguishing between the two classes, e.g. all blue rectangles belong to the positive class\cite{van2016massively}. Concept drift can be introduced by changing the classification rule. 

\textbf{SEA Generator} generates 3 numerical attributes, that vary from 0 to 10, where only 2 of them are relevant to the classification task. A classification function is chosen, among four possible ones. These functions compare the sum of the two relevant attributes with a threshold value, unique for each of the classification functions\cite{street2001streaming}. Depending on the comparison the generator will classify an instance as one of the two possible labels. Abrupt concept drift is introduced by changing the classification function.

\textbf{LED Generator} generates samples by emulating the elements of a 7-segment display. The goal is to predict the digit displayed on a seven-segment LED display, where each attribute has a 10\% chance of being inverted. Additional features are added in order to generate noise and concept drift \cite{breiman1984classification}.

\textbf{Hyperplane Generator} generates a binary classification problem of determining if a point is below or under a rotating hyperplane. Hyperplanes are useful for simulating time-changing concepts because the orientation and position of the hyperplane can change in a continuous manner by changing the relative size of the weights \cite{hulten2001mining}. Drift is introduced by changing the weights and reversing the direction of the rotation.
    
\textbf{Random Radial Basis Function (RBF) Generator} generates a number of centroids. Each has a random position in a Euclidean space, standard deviation, weight and class label. Each example is defined by its coordinates in Euclidean Space and a class label referring to a centroid close by.  Centroids move at a certain speed, generating gradual concept drift \cite{van2016massively}.

\textbf{covtype} contains the forest cover type for 30 x 30 meter cells obtained from US Forest Service (USFS). It contains 581012 instances and 54 attributes, and it has been used in several papers on data stream classification. The source of this dataset, as well as the elec and pokerhand, is the website\footnote{\url{https://moa.cms.waikato.ac.nz/datasets}} of the popular online learning framework MOA.

\textbf{elec} contains data was collected from the Australian New South Wales Electricity Market. In this market, prices are not fixed and are affected by the demand and supply of the market. They are set every five minutes. The class label identifies the change of the price relative to a moving average of the last 24 hours.

\textbf{pokerhand.} Each record of the dataset is an example of a hand consisting of five playing cards drawn from a standard deck of 52. Each card is described using two attributes (suit and rank), for a total of 10 predictive attributes. Their class attribute describes the “Poker Hand”.

\section{Experiments and results}
\label{experiments}

The experimental setup of this research aims to provide reproducibility and allow future research to use the same experimental framework. However, the current experimental machine learning tooling is mainly divided between Java-based and Python-based implementations. Some researchers implement their experiments using tools built around Weka\footnote{\url{https://www.cs.waikato.ac.nz/ml/weka/}} \cite{holmes1994weka}: the AutoML framework AutoWeka \cite{thornton2013autoweka} or the data stream mining framework MOA \cite{bifet2010moa}. Others prefer using solutions from the scikit-learn\footnote{\url{https://scikit-learn.org/}} environment: the AutoML framework auto-sklearn \cite{feurer2015autosklearn} or the multi-output streaming platform scikit-multiflow\footnote{\url{https://scikit-multiflow.github.io/}} \cite{montiel2018scikit}. Furthermore, some of these tools rely on specific versions of operating-system-wide dependencies. To overcome these problems, an experimental setup using Docker\footnote{\url{https://www.docker.com/}} containers is used. Each container provides an isolated environment that includes all the dependencies required by a certain tool or framework. This way, the experimental setup is consistent across environments and scalable. New experiments using a new tool or framework can be added by simply adding a new container. Moreover, the same containers used for research and development can then be used in production, ensuring parity between development and production environments.

For providing a stream of data similar to the ones usually used in production, Apache Kafka\footnote{\url{https://kafka.apache.org}}, a widely-adopted distributed and scalable \cite{kreps2011kafka} Pub/Sub broker was used. The datasets described in Table \ref{table:datasets} were streamed as topics to the broker and replayed during training and testing. 

Additionally, to implement all the experiment types described in Section \ref{methods}, developing new scripts and techniques was required. The resulting implementations are collected under a Python package called \textbf{automl-streams}\footnote{\url{https://pypi.org/project/automl-streams}} available on GitHub\footnote{\url{https://github.com/AlexImb/automl-streams}}. It is built around the scikit-multiflow framework and its components are therefore cross-compatible with scikit-learn and scikit-multiflow.

For now, the automl-streams package includes:

\begin{itemize}
    \item KafkaStream, a way to create a stream for scikit-mutliflow and other Python libraries from a Kafka topic.
    \item EvaluatePretrained, an evaluator for scoring pretrained models against a stream.
    \item implementations of classifiers using techniques from MetaStream \cite{rossi2014metastream} and BLAST \cite{van2018online}.
    \item other experimental classifiers that use meta-learning and meta-feature extraction.
    \item helper functions for publishing CSV files, Pandas DataFrames and OpenML Datasets to Kafka topics.
\end{itemize}

For the rest of the section experiments and results that fit in one of the experiment-type categories explained above will be presented.

\subsection{Classic model}

For this experiment, the batch training algorithms described in Table \ref{table:batch_algos} are used. The parameters of these classifiers are the default ones from their corresponding implementation in sklearn. 

\begin{table}[h]
\centering
\renewcommand{\arraystretch}{1.25}
\begin{tabular}{|l|l|} \hline
\textbf{Algorithm} & \textbf{Implementation}\\ \hline
RandomForestClassifier & sklearn.ensemble \\ \hline
DecisionTreeClassifier & sklearn.tree \\ \hline
KNeighborsClassifier & sklearn.neighbors \\ \hline
LinearSVC & sklearn.svm \\ \hline
\end{tabular}
\caption{Classifiers used for batch training}
\label{table:batch_algos}
\end{table}

\begin{figure}[h]
\centering 
\epsfig{file=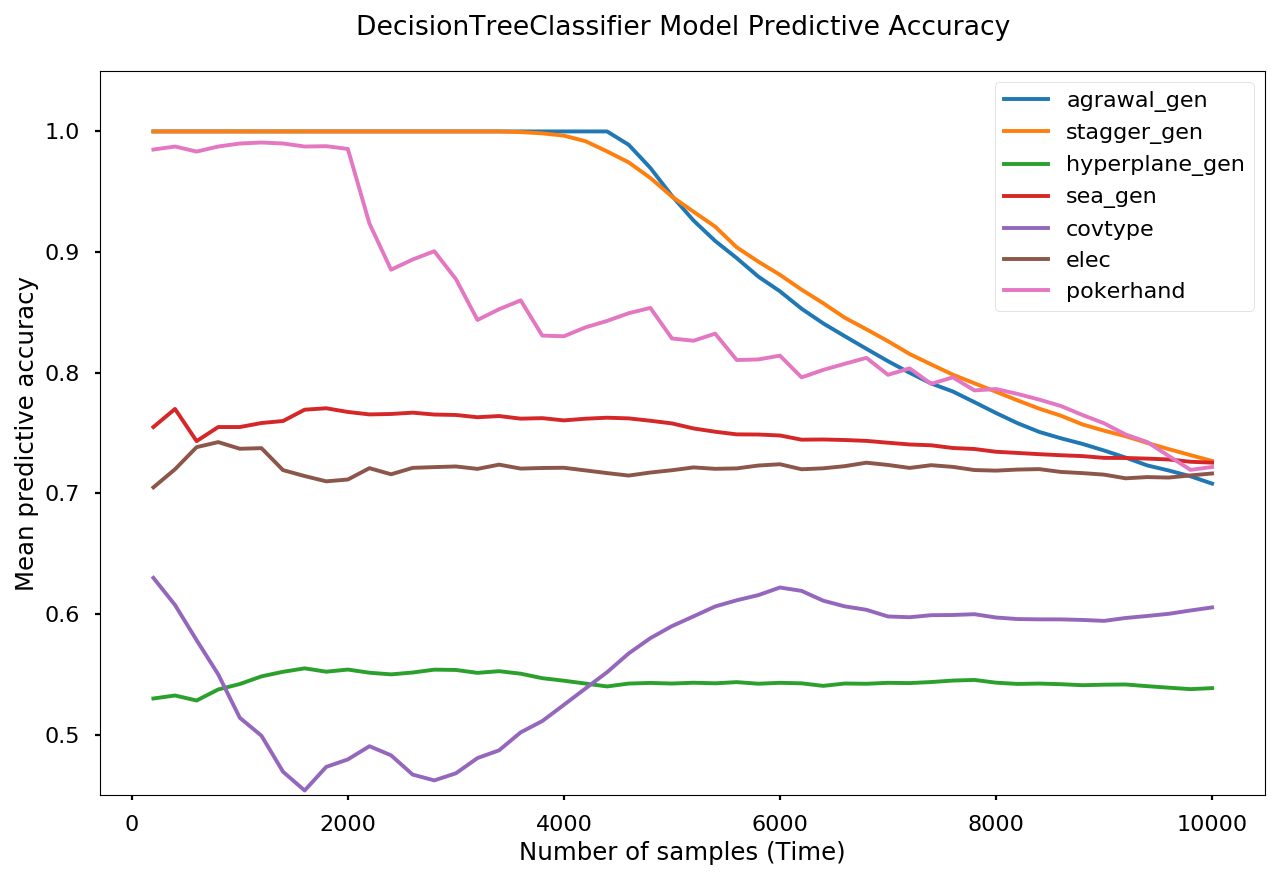, width=7.8cm}
\caption{Batch-trained model accuracy over time for a subset of datasets}
\label{fig:batchdt}
\end{figure}

The predictive accuracy and Cohen's kappa metrics were measured for all datasets. Figure \ref{fig:batchdt} shows a plot of the predictive accuracy for a single classifier and a selection of datasets. Figure \ref{fig:batchall}, showing the predictive accuracy of all classifiers and all datasets, can be found in Appendix \ref{appendix:a}.

\begin{figure}[h]
\centering 
\epsfig{file=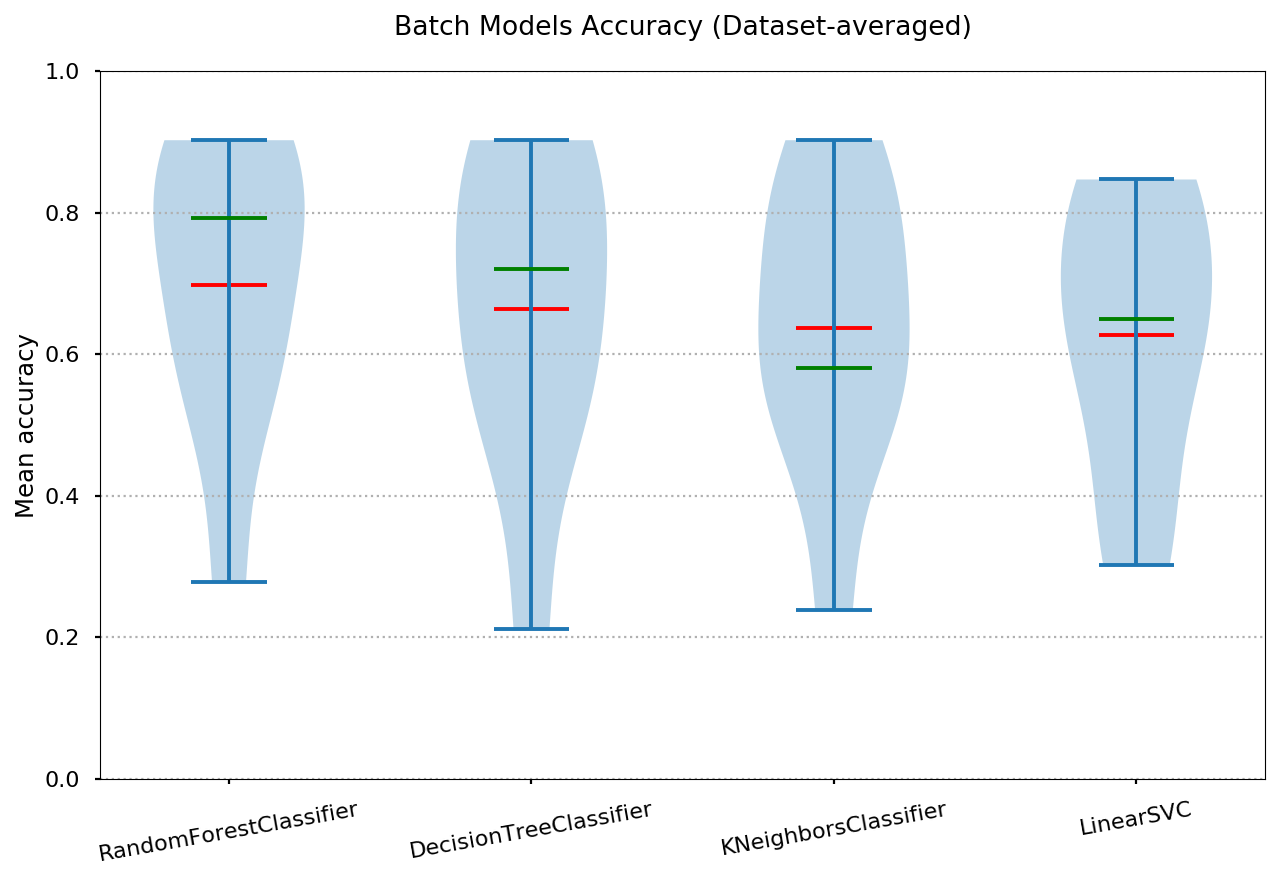, width=8cm}
\caption{Batch-trained models accuracy averages}
\label{fig:batch_violins}
\end{figure}

A comparison of the selected algorithms' average accuracy over all datasets is shown in Figure \ref{fig:batch_violins}. Each violin plot shows the distribution of the average accuracy across all datasets, the mean (red), and the median (green).

\vspace{3cm}

\subsection{Online learning model}

For online learning, the algorithms listed in Table \ref{table:online_algos} are used. Some of them incorporate explicit drift detection techniques. The evaluation method used for measuring is the prequential method described in Subsection \ref{evaluation}.

\begin{table*}[h]
\centering
\renewcommand{\arraystretch}{1.25}
\begin{tabular}{|l|l|l|l|} \hline
\textbf{Algorithm} & \textbf{Implementation} & \textbf{Type}  & \textbf{Drift detection} \\ \hline
HoeffdingTree & skmultiflow.trees &  Instance incremental & None \\ \hline
KNearestNeighbors (KNN) & skmultiflow.lazy & Batch incremental & None \\ \hline
PerceptronMask & skmultiflow.neural\_networks & Instance incremental & None \\ \hline
SGDClassifier & sklearn.linear\_model & Instance incremental &  None \\ \hline
HoeffdingAdaptiveTree (HAT) & skmultiflow.trees & Batch incremental & ADWIN \\ \hline
LeverageBagging  & skmultiflow.meta & Ensemble &  ADWIN \\ \hline
OzaBaggingAdwin & skmultiflow.meta & Ensemble & ADWIN \\ \hline
\end{tabular}
\caption{Classifiers used for online learning}
\label{table:online_algos}
\end{table*}

\begin{figure}[h!]
\centering 
\epsfig{file=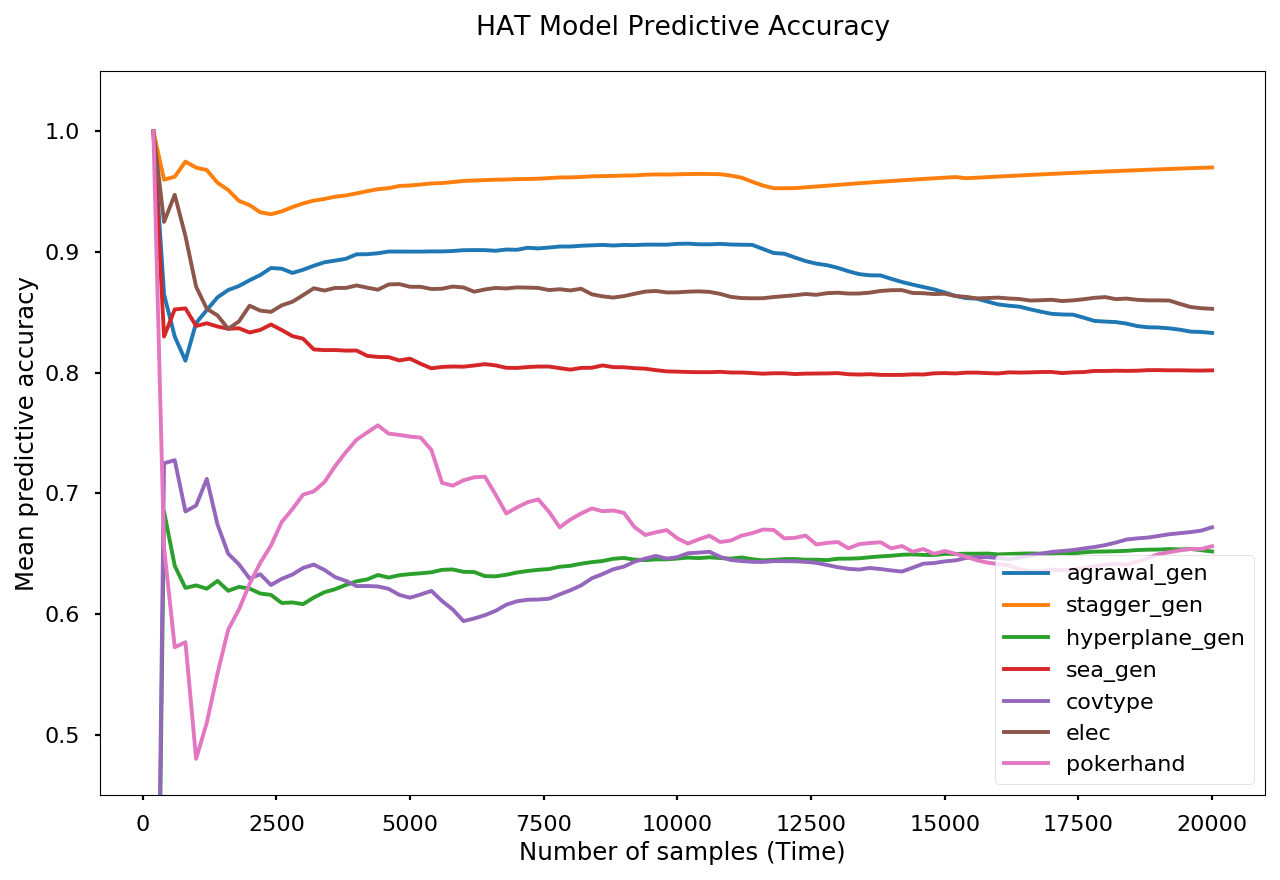, width=8cm}
\caption{Online models without and with (HAT) drift detection. Accuracy over time for a subset of datasets}
\label{fig:online_hat}
\end{figure}

Figure \ref{fig:online_hat} shows how the mean predictive accuracy changes over time for a single online classifier and a selected subset of datasets. Appendix \ref{appendix:a} contains Figure \ref{fig:online_all} that includes all the datasets and algorithms.

\begin{figure}[h]
\centering 
\epsfig{file=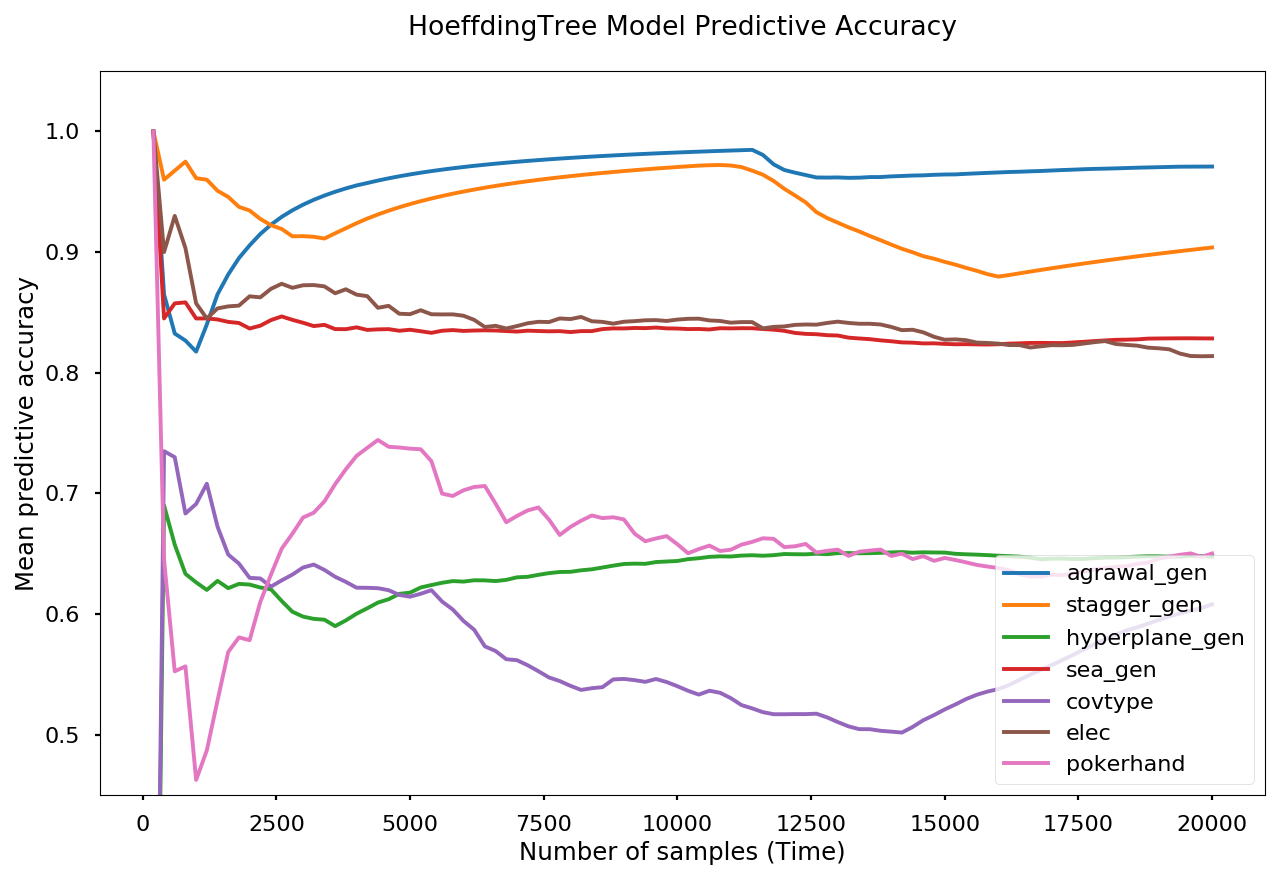, width=8cm}
\end{figure}

A violin plot, showing the distribution of the average accuracy across topics for each of the online learning algorithms is depicted in Figure \ref{fig:online_violin} from Appendix \ref{appendix:a}.

\subsection{Pretrained AutoML model}

\begin{figure}[h!]
\centering 
\epsfig{file=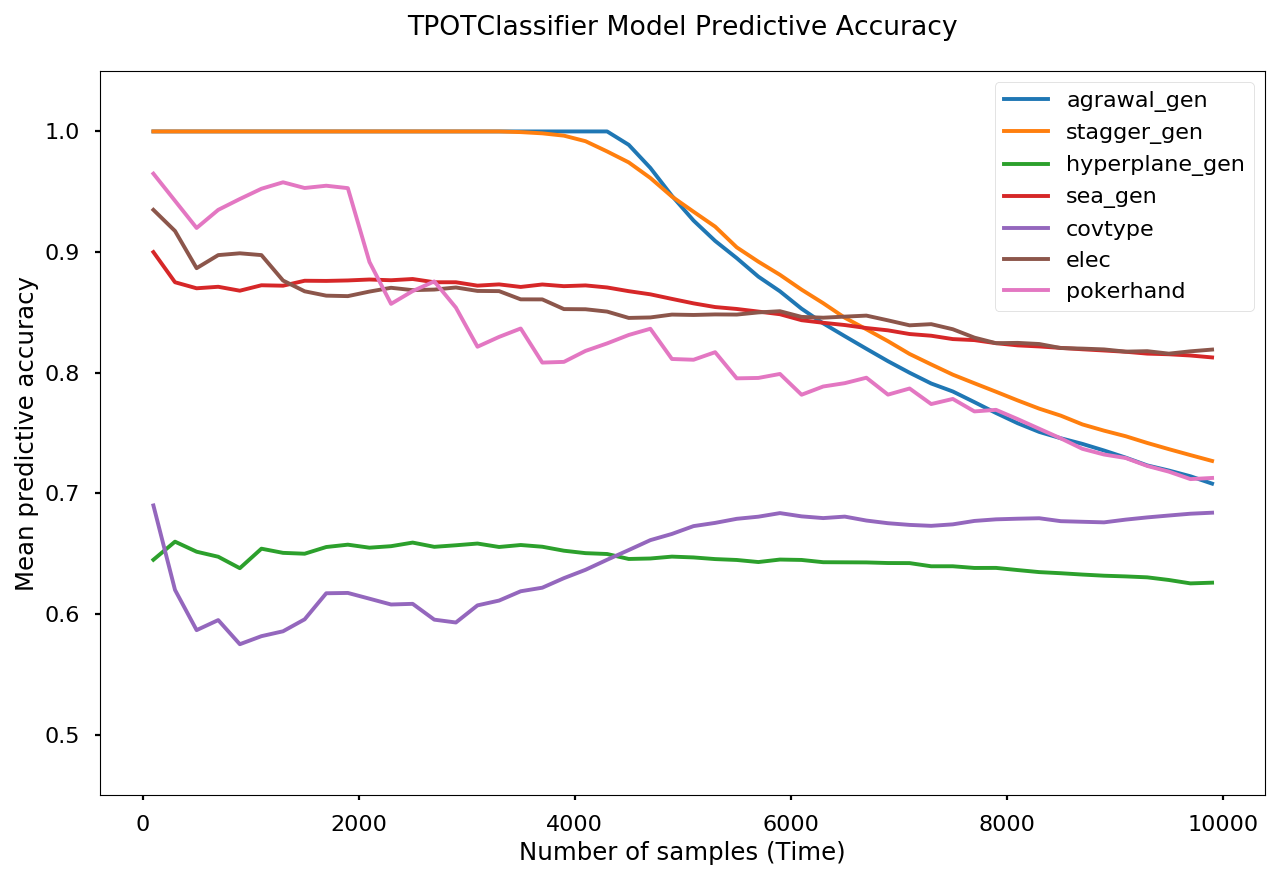, width=8cm}
\caption{AutoML models accuracy over time for a subset of datasets}
\label{fig:automl_tpot}
\end{figure}

For automatically generating models, two popular \cite{gijsbers2019open} AutoML frameworks were selected: auto-sklearn and TPOT. Each framework has a budget of 180 seconds for training. The resulting model is evaluated for the streaming data. TPOT is restricted to maximum 5 generations (generations=5) and a maximum population of 20 instances for each generation (population\_size=20). For auto-sklearn, a maximum time for fitting each model is set to 30 seconds (per\_run\_time\_limit=30). Example executions of this experiment for both frameworks are depicted in Figure \ref{fig:automl_tpot}.

\begin{figure}[h]
\centering 
\epsfig{file=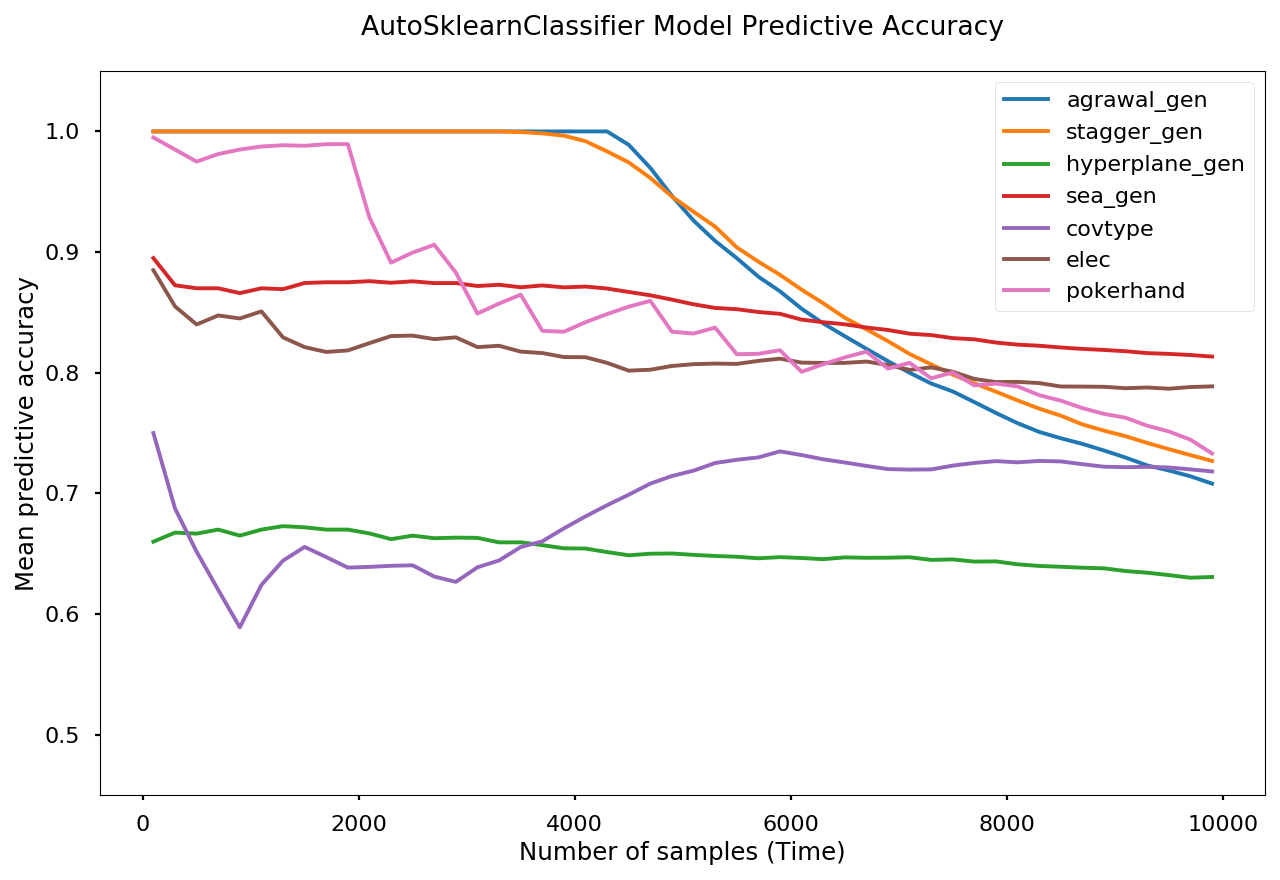, width=8cm}
\end{figure}

A violin plot, showing the distribution of the average accuracy across topics for each of the AutoML-generated models is depicted in Figure \ref{fig:automl_violin} from Appendix \ref{appendix:a}.

\vspace{1cm}
\subsection{Online meta-learning model}

The meta-learning algorithm implemented is inspired by the solution proposed by Rossi et al., MetaStream \cite{rossi2014metastream}, but adapted to solve classification problems. Another difference is that the set of base-estimators can contain more than two estimators. For this experiment, the following base-estimator algorithms are used: HoeffdingTree, KNN, PerceptronMask, SGDClassifier. The meta-learner responsible for choosing the active estimator and incorporating the meta-knowledge is selected to be an instance of a sklearn.linear\_model.SGDClassifier. For extracting meta-features the pymfe\footnote{\url{https://pypi.org/project/pymfe}} library is used. The following meta-feature categories are extracted: general, statistical, and info-theory. A description of which meta-features are included in each category is available on the pyfme website\footnote{\url{https://pymfe.readthedocs.io/en/latest/auto_pages/meta_features_description.html}}. In total, 45 features are extracted for each tumbling window of 300 samples. These meta-features and the index of the predictor with the best accuracy for the current window are used for training the meta-learner online. This way, the meta-knowledge extracted from the meta-feature is gradually incorporated into the meta-model. At the end of each window, the meta-learner predicts which base-learner will be used for making predictions during the next window. 

\begin{figure}[h]
\centering 
\epsfig{file=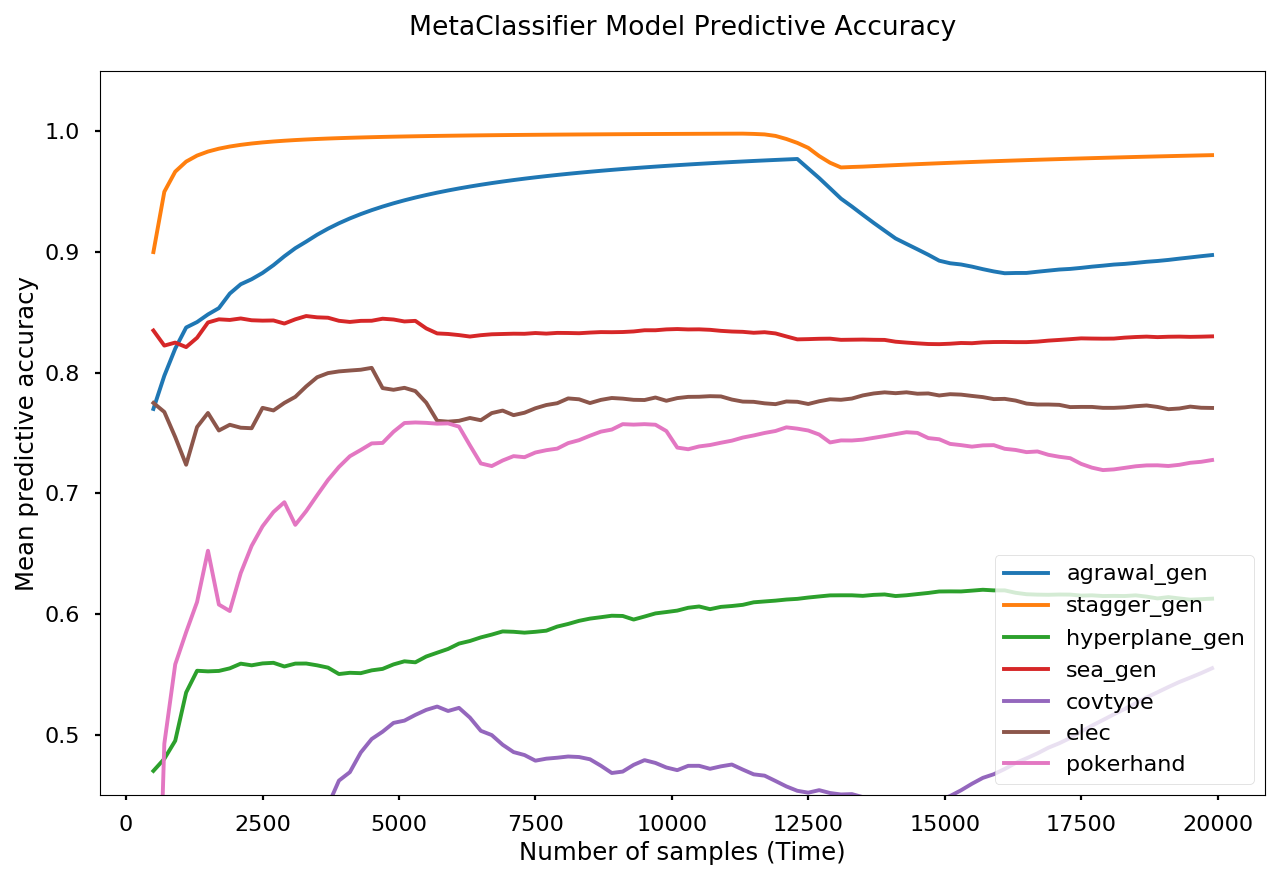, width=8cm}
\caption{Meta-learning model accuracy over time for a subset of datasets}
\label{fig:meta}
\end{figure}

For comparison, an implementation that only selects the predictor for the next window to be the one that performed best in the current window is implemented. The meta-knowledge is not considered to show the importance of the meta-features in the MetaClassifier implementation. This "algorithm" is called LastBestClassifier and a comparison to the MetaClassifier is in Figure \ref{fig:meta_violin} from Appendix \ref{appendix:a}.

\begin{figure*}[hbt!]
\centering 
\epsfig{file=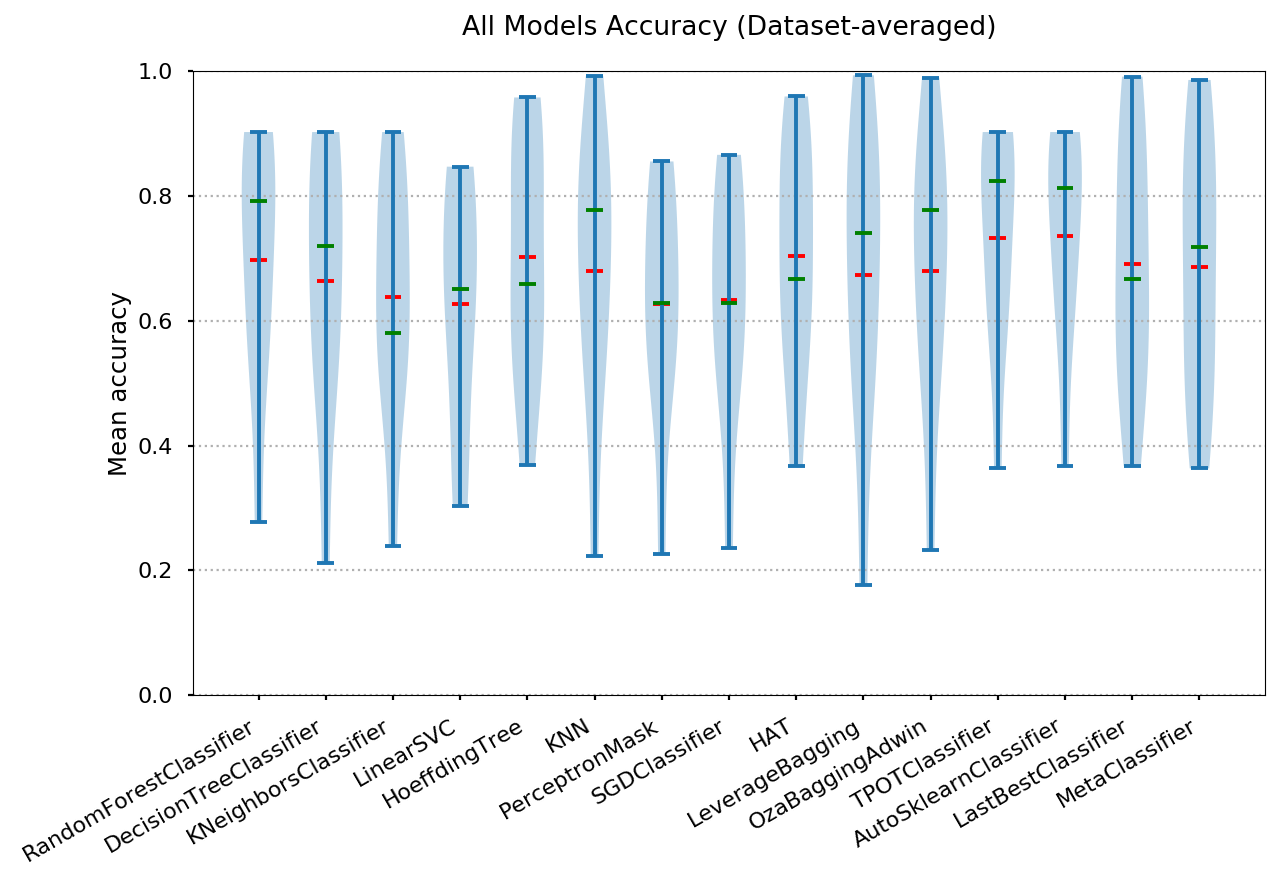, width=16cm}
\caption{All models accuracy averages distribution, mean (red) and median (green)}
\label{fig:all_violins}
\end{figure*}

\section{Discussion}

The experiments performed show that the presence of concept drift in streaming data is an important factor that leads to a decrease in predictive accuracy over time. In the case of the batch-trained models from the first experiment, the decrease is significant for both synthetic data with abrupt drift and real data. The online learning algorithms performed consistently better in the presence of both abrupt and slow-changing drift. The online learning algorithms that include explicit drift detection techniques (e.g. HAT that includes ADWIN) scored the highest and the most consistent accuracy over time among all selected algorithms.

Furthermore, the AutoML frameworks generated models with a higher initial predictive accuracy but it decreased rapidly when encountering concept drift. The meta-learning approach of a heterogeneous ensemble of online learners resulted in better performance compared to some of the online ones alone but worse than the online ensembles equipped with ADWIN. The improvement of using the selected categories of meta-features led only to marginal improvements compared to selecting the best classifier of the last window. These findings are in line with the claims from the literature saying that meta-feature extracted from the stream lead to marginal improvements at a high computational cost \cite{van2018online}. 

However, having a heterogeneous ensemble of diverse algorithms and selecting an active one online proved to be a useful technique for data streams. As shown in Figure \ref{fig:all_violins}, the distribution of the accuracy averages for the last type of algorithms has its minimum and maximum values close to the best values of all the other algorithms. In combination with HPO and better meta-knowledge, it can be a valuable addition to an AutoML tool for streaming data. 

\section{Conclusion}

This research surveyed and summarised the existing AutoML and online learning tools and techniques applicable to streaming data. It proposed a scalable and portable benchmarking framework based on Apache Kafka streams and Docker containers. Consequently, using the introduced framework, it empirically showed how the predictive performance of models is influenced by concept drift over time. Finally, a meta-learning technique based on meta-features extraction was introduced and benchmarked. The results showed that the implemented meta-learning approach provides marginal improvements compared to state-of-the-art ensembles equipped with drift detection, but it is useful for selecting the active estimator in a heterogeneous ensemble.  Among other techniques, it has the potential to be the core of an AutoML tool for data streams.

\section{Future Work}

Considering the importance and the diversification of the AutoML problem for data streams, future work ought to be performed to empirically cover other existing techniques and develop new ones. Using automl-streams, the experimental framework proposed in this work, more algorithms can be implemented and easily benchmarked in the same way used in this research. Furthermore, automl-streams can be used to plot other metrics such as training and prediction times for a better assessment of the computational cost of the presented techniques. Moreover, new AutoML frameworks, including the Java-based ones (H2O.ai, AutoWeka, etc.), can be added and compared to the ones used above. Finally, considering the need for an end-to-end AutoML framework in the industry, automl-streams may be extended with Auto Cleaning, AutoFE, and HPO components.

\bibliographystyle{abbrv}
\bibliography{automl}  
%

\newpage

\appendix
\section{DETAILED AND ADDITIONAL PLOTS}
\label{appendix:a}

Includes plots for all the datasets and additional plots for the online, AutoML and meta-learning experiments.

\begin{figure}[h!]
\centering 
\epsfig{file=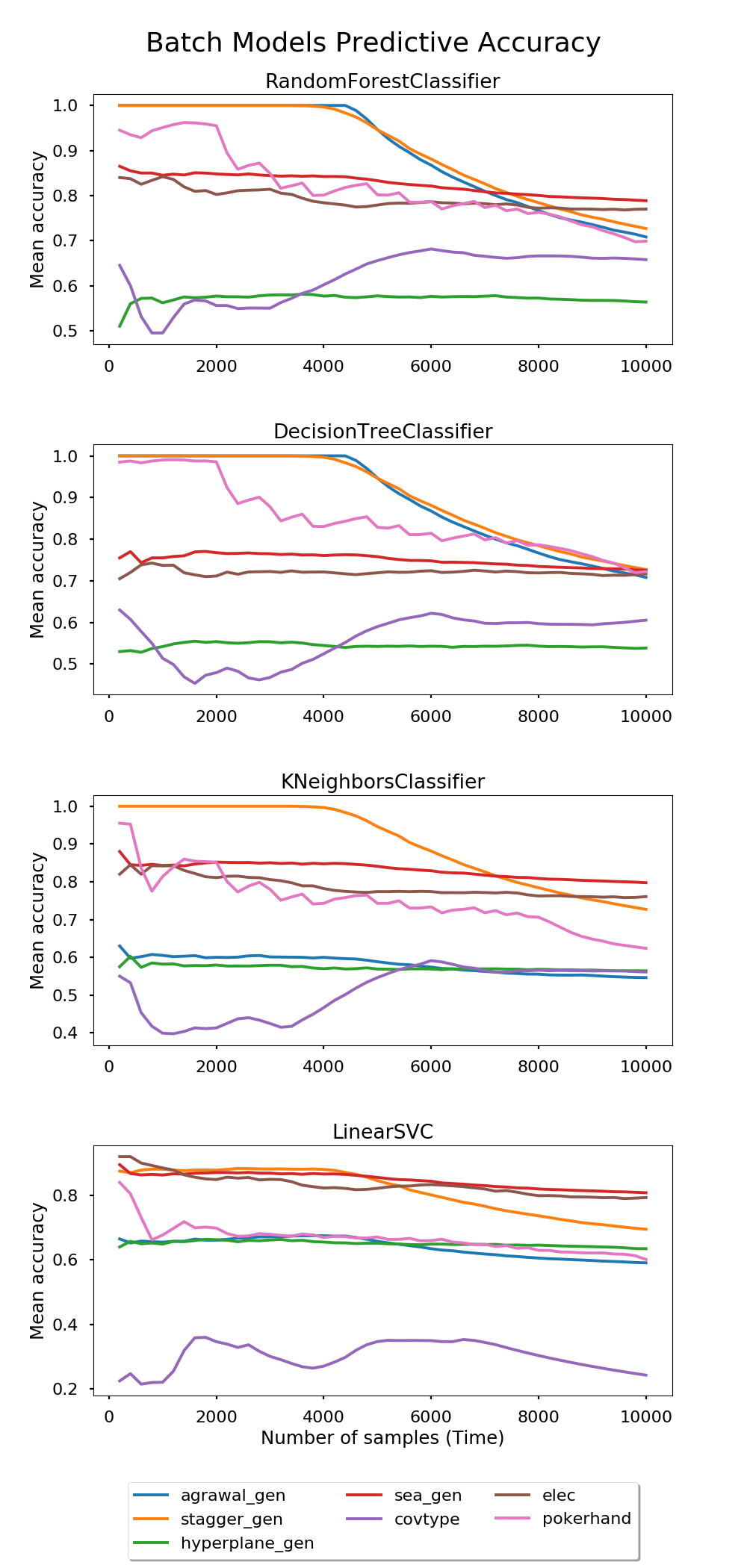, width=8cm}
\caption{Detailed plots of batch models accuracy over time grouped by algorithm}
\label{fig:batchall}
\end{figure}

\begin{figure}[h!]
\centering 
\epsfig{file=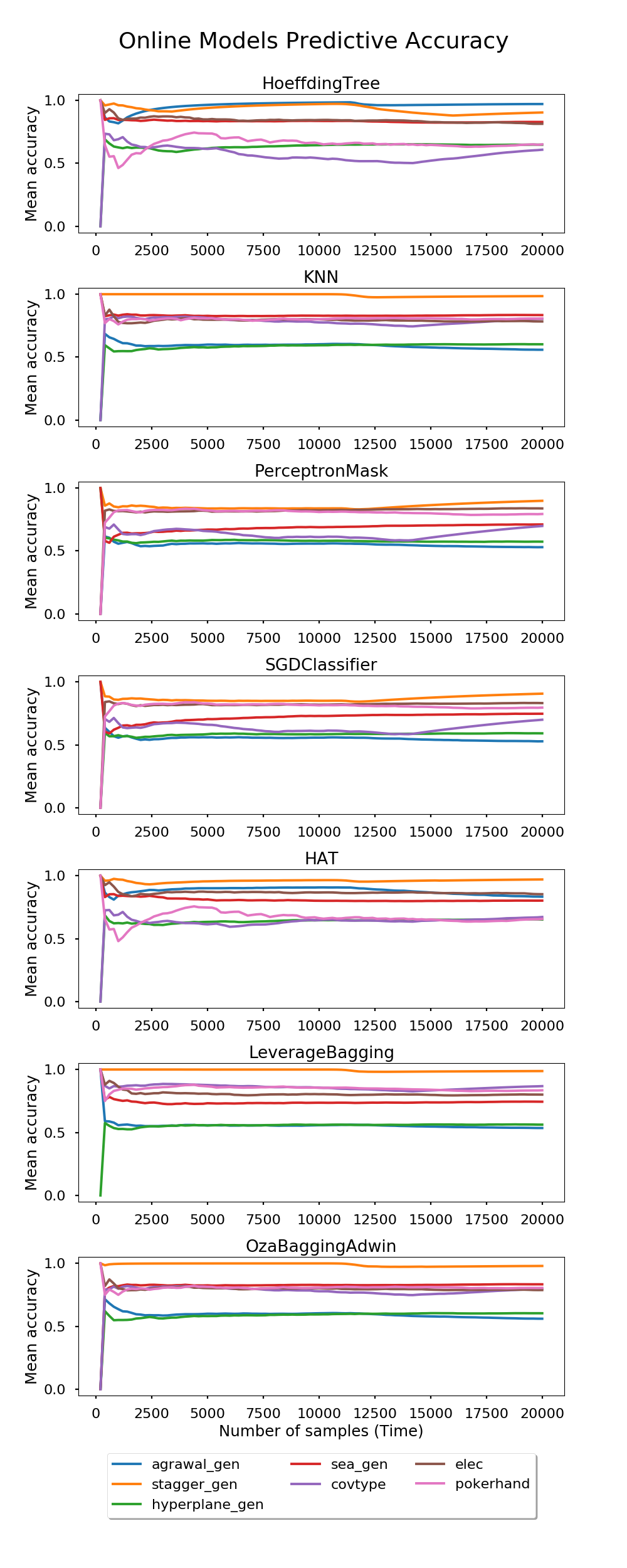, width=8cm}
\caption{Detailed plots of online models accuracy over time grouped by algorithm}
\label{fig:online_all}
\end{figure}

\begin{figure}[h!]
\centering 
\epsfig{file=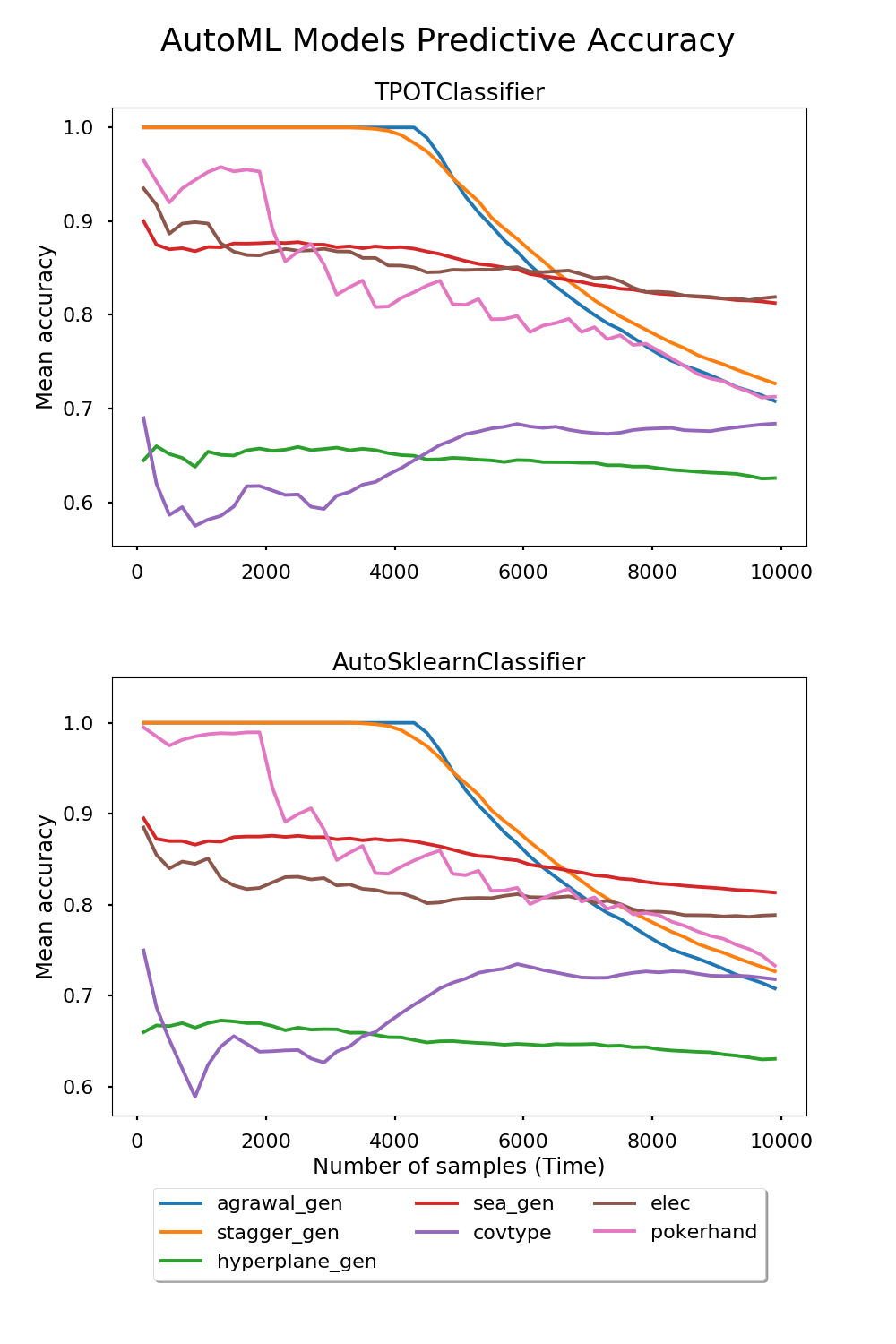, width=7.5cm}
\caption{Detailed plots of AutoML models accuracy over time grouped by algorithm}
\label{fig:automlall}
\end{figure}

\begin{figure}[h!]
\centering 
\epsfig{file=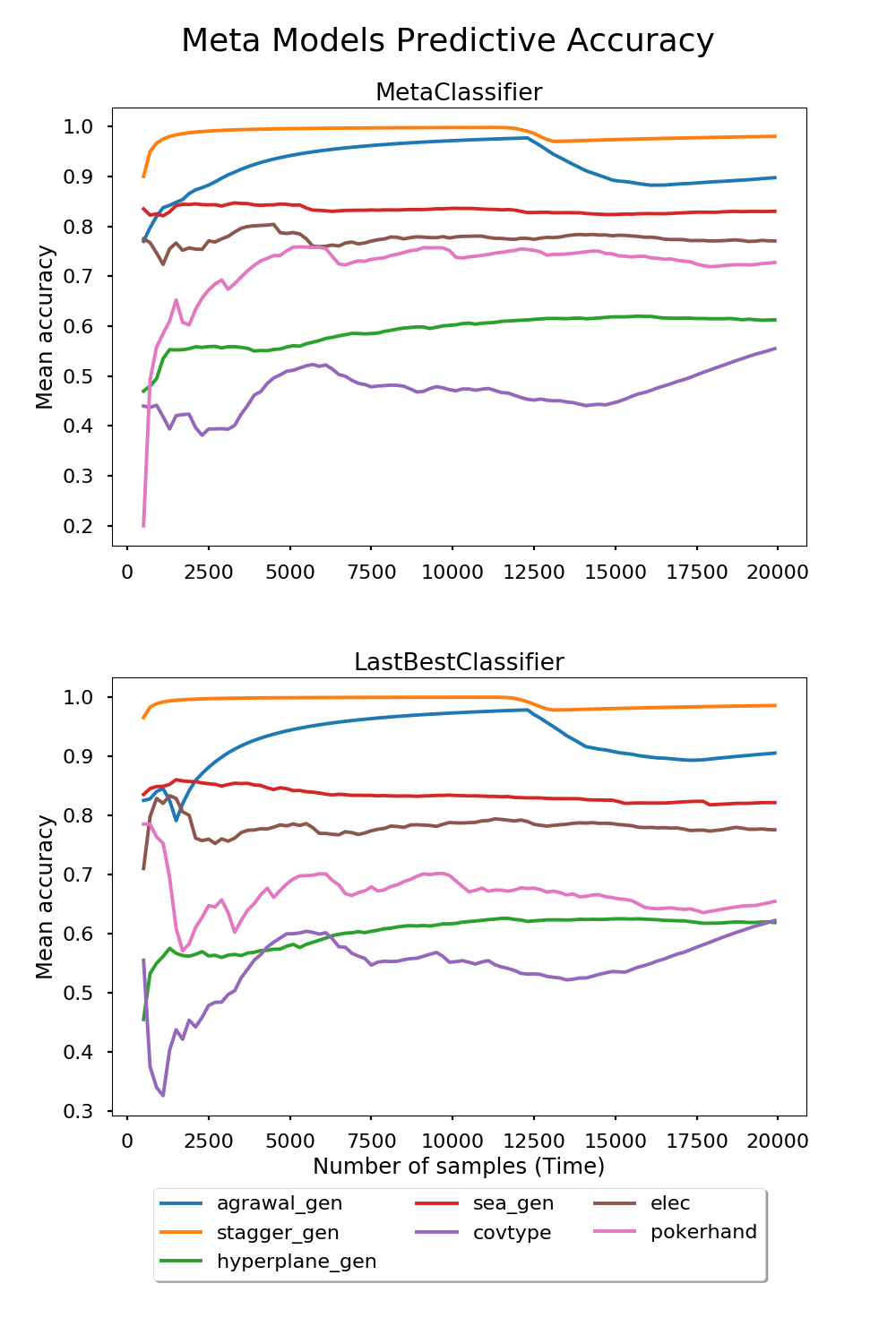, width=7.5cm}
\caption{Detailed plots of meta models accuracy over time grouped by algorithm}
\label{fig:metaall}
\end{figure}

\begin{figure}[h!]
\centering 
\epsfig{file=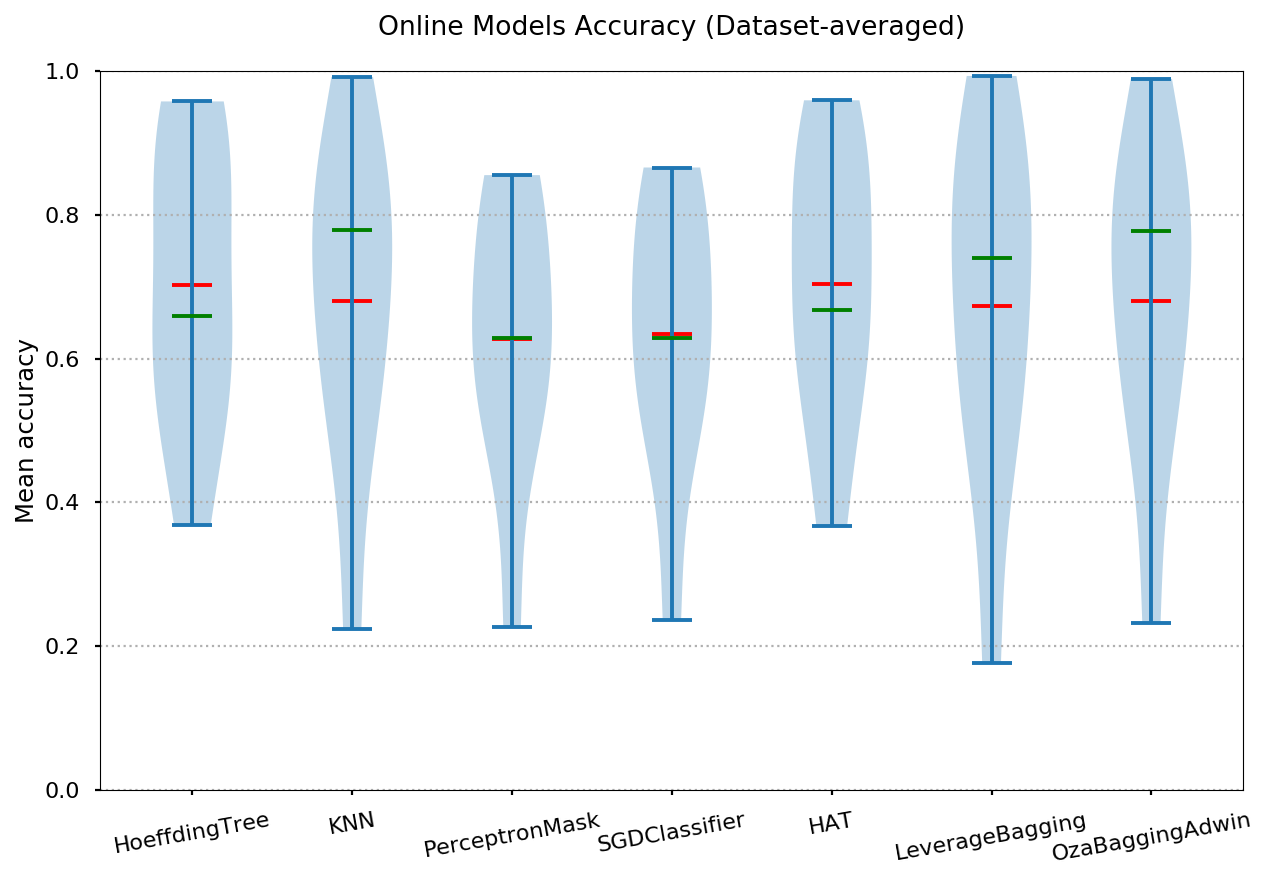, width=8cm}
\caption{Online models accuracy averages}
\label{fig:online_violin}
\end{figure}

\begin{figure}[h!]
\centering 
\epsfig{file=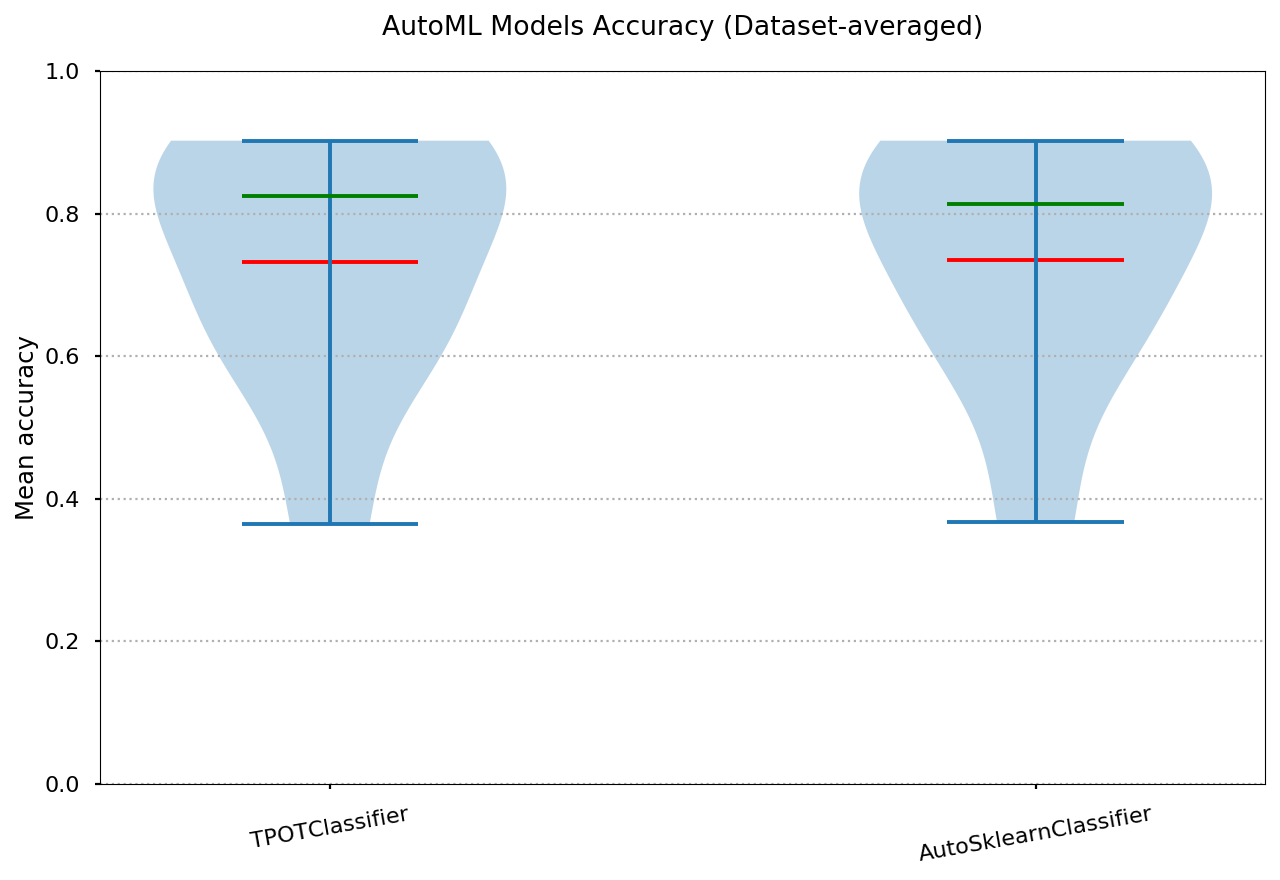, width=8cm}
\caption{AutoML models accuracy averages}
\label{fig:automl_violin}
\end{figure}

\begin{figure}[h!]
\centering 
\epsfig{file=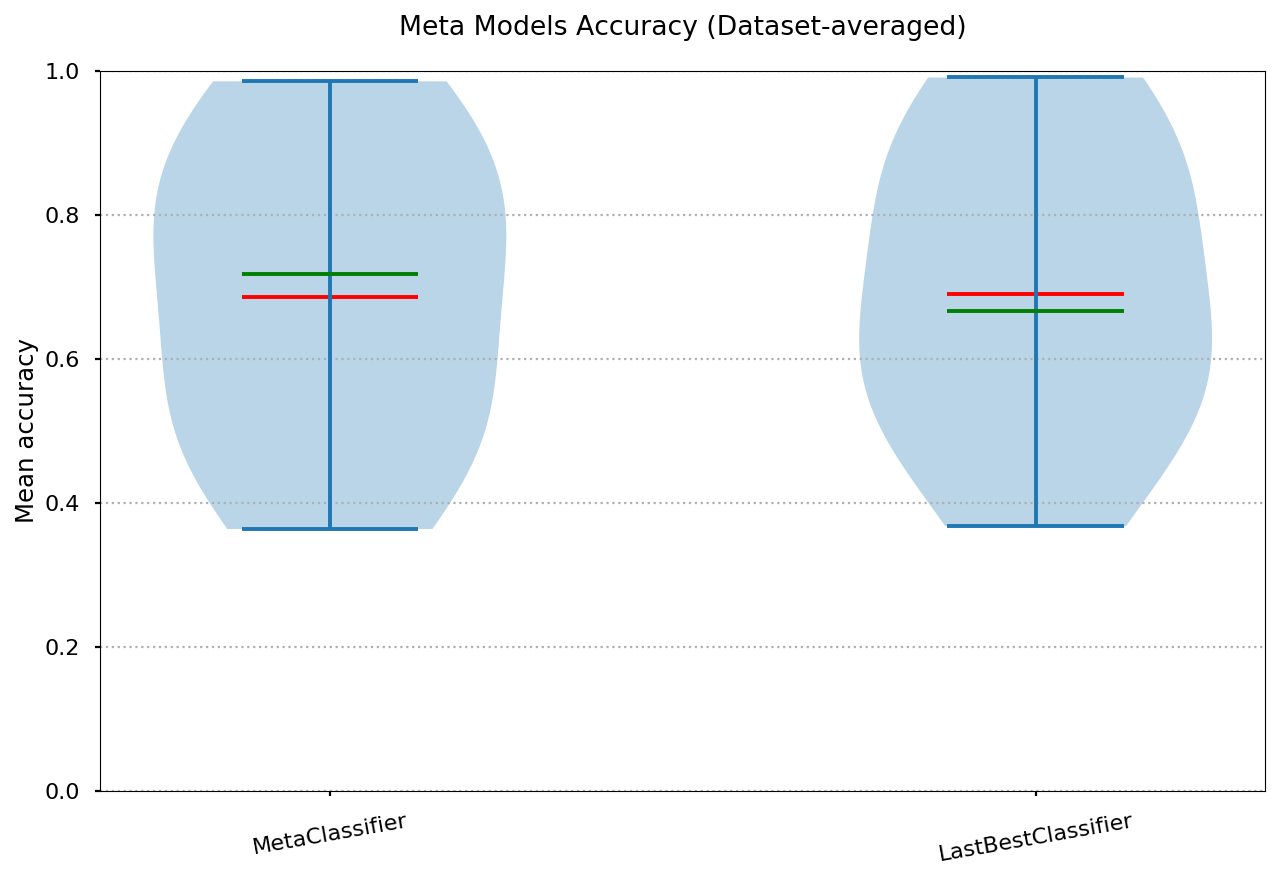, width=8cm}
\caption{Meta-learning models accuracy averages}
\label{fig:meta_violin}
\end{figure}

\end{document}